\documentclass[runningheads]{llncs}

\usepackage{eccv}

\usepackage{eccvabbrv}

\usepackage{graphicx}
\usepackage{booktabs}
\usepackage{makecell}
\usepackage{overpic}
\usepackage{color}
\usepackage{indentfirst}
\usepackage{epsfig}
\usepackage{amsmath}
\usepackage{amssymb}
\usepackage{multirow}
\usepackage{array}
\usepackage{booktabs}
\usepackage{bm}

\usepackage[accsupp]{axessibility}  

\usepackage{hyperref}

\usepackage{orcidlink}

\begin{document}

\title{VirtualModel: Generating Object-ID-retentive Human-object Interaction Image by Diffusion Model for E-commerce Marketing} 

\titlerunning{Abbreviated paper title}

\author{Binghui Chen \and
Chongyang Zhong \and Wangmeng Xiang \and Yifeng Geng \and
Xuansong Xie}


\institute{Institute for Intelligent Computing, Alibaba Group \\
\email{chenbinghui@bupt.cn, zhongchongyang.zzy@alibaba-inc.com, \\marquezxm@gmail.com, cangyu.gyf@alibaba-inc.com, xingtong.xxs@taobao.com}\\
\href{https://aigcdesigngroup.github.io/replace-anything/}{Project:https://aigcdesigngroup.github.io/replace-anything}}

\maketitle

\begin{abstract}
  Due to the significant advances in large-scale text-to-image generation by diffusion model (DM), controllable human image generation has been attracting much attention recently. Existing works, such as Controlnet \cite{zhang2023adding}, T2I-adapter \cite{mou2023t2i} and HumanSD \cite{ju2023humansd} have demonstrated good abilities in generating human images based on pose conditions, they still fail to meet the requirements of real e-commerce scenarios. These include (1) the interaction between the shown product and human should be considered, (2) human parts like face/hand/arm/foot and the interaction between human model and product should be hyper-realistic, and (3) the identity of the product shown in advertising should be exactly consistent with the product itself. To this end, in this paper, we first define a new human image generation task for e-commerce marketing, i.e., \textbf{O}bject-ID-retentive \textbf{H}uman-object Interaction image \textbf{G}eneration (OHG), and then propose a \textbf{VirtualModel} framework to generate human images for product shown, which supports displays of any categories of products and any types of human-object interaction. As shown in Figure \ref{fig_intro}, VirtualModel not only outperforms other methods in terms of accurate pose control and image quality but also allows for the display of user-specified product objects by maintaining the product-ID consistency and enhancing the plausibility of human-object interaction. Codes and data will be released.
\end{abstract}

\section{Introduction}
\label{sec:intro}

Recently, the field of generative models has witnessed significant progress with the diffusion model (DM) \cite{ho2020denoising,song2020score,rombach2022high,ho2022classifier} emerging as a new paradigm for hyper-realistic image synthesis. It has become a popular and powerful architecture in Generative AI\cite{dhariwal2021diffusion}. Diffusion model actually and really gives chances of dropping the image synthesis into our real life, such as creative imagining by text-to-image \cite{rombach2022high,ramesh2022hierarchical,nichol2021glide}, image colorization/restoration \cite{saharia2022palette}, image super-resolution \cite{saharia2022image}, virtual try-on \cite{zhu2023tryondiffusion}, image editing \cite{yang2023paint,couairon2022diffedit,mokady2023null}, customized image generation \cite{ruiz2023dreambooth,shi2023instantbooth} and so on.

It is worth noting that most of the aforementioned methods primarily focus on generating impressive natural landscapes or animal images, rather than hyper-realistic human images. To address this limitation, several controllable diffusion models have been proposed, namely ControlNet \cite{zhang2023adding}, T2I-Adapter \cite{mou2023t2i} and HumanSD \cite{ju2023humansd}. These models utilize keypoints annotations of the human body to guide the generation of corresponding human parts, thereby enhancing the realism and plausibility of the generated human images.

\begin{figure}[t]
  \centering
  \includegraphics[width=1.0\linewidth]{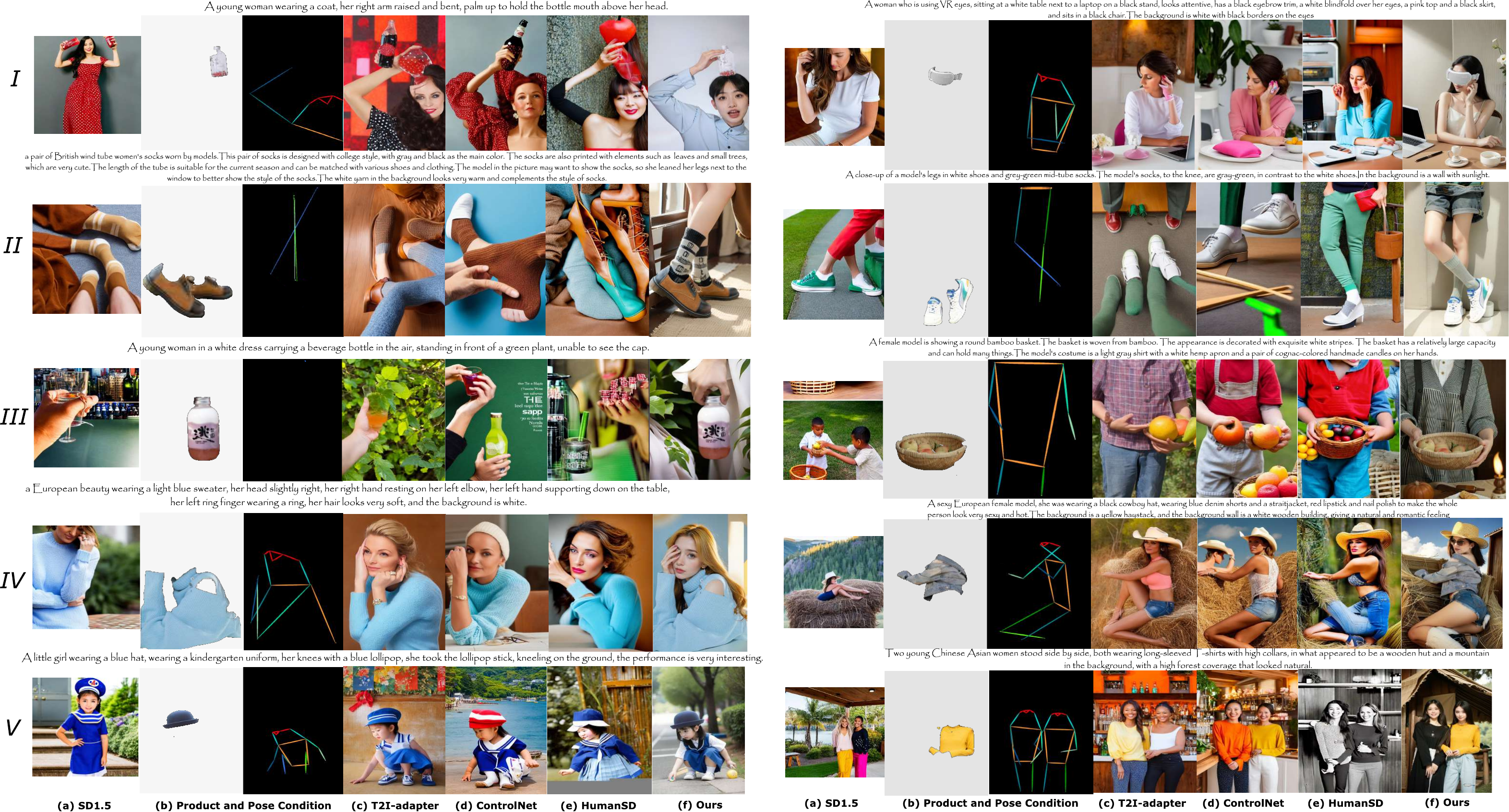}
  \vspace{-1em}
  \caption{\textbf{Example Results and Visual Comparisons.} This paper mainly focuses on \textbf{O}bject-ID-retentive \textbf{H}uman-object Interaction image \textbf{G}eneration (OHG) task for E-commerce marketing scenario. Each row contains: (a) a generation by text-guided Stable Diffusion \cite{rombach2022high}, (b) given product and pose conditions for OHG, (c) a generation by T2I-adapter \cite{mou2023t2i}, (d) a generation by ControlNet \cite{zhang2023adding}, (e) a generation by HumanSD \cite{ju2023humansd} and (e) a generation by our proposed \textit{VirtualModel}. Since (c,d,e) do not support OHG task, only pose conditions are used. Comparing with other methods, when given the target products and the corresponding pose-skelenton images, \textit{VirtualModel} can generate hyper-realistic marketing images in terms of reasonability of human-object interaction, image quality and challenging local-poses. \textbf{Best viewed with zoom-in}.}\label{fig_intro}
  \vspace{-1.5em}
\end{figure}

However, in actual, as shown in Figure. \ref{fig_intro}.(a,c,d,e), these above controllable human generation methods still suffers from bad image quality, inaccurate pose prediction and suboptimal or unreasonable human-object interaction, letting alone interacting with other specified objects. As a result, these models have yet to fully meet the requirements of certain real-life applications, such as e-commerce marketing where the generated human should be hyper-realistic, human can interact with the user-specified product and, different types of pose should be supported.

To this end, the aim of this paper is for the first time to really push the controllable human image generation technology into real-world application, \ie, e-commerce marketing. Specifically, we first define a new human image generation task, called \textbf{O}bject-ID-retentive \textbf{H}uman-object Interaction image \textbf{G}eneration (OHG). In this task, the inputs consist of the user-specified product, the corresponding human-pose and the text description, while the outputs are the high-quality images depicting the human showcasing the specified product. Under this definition, we collect and build a high-quality human-object interacted dataset, named \textit{HoIHuman}, which consists of 3M professional e-commerce marketing images with high quality, high resolution, diverse layouts, rich human-object interactions and wide ranges of types of products. It has comprehensive annotations, such as the fine-grained whole-body skeletons (133 keypoints), the fine-grained object masks and canny-edges, as well as the high-level image captions and attributes. Next, we propose a \textbf{\textit{VirtualModel}} framework for this OHG task. It is mainly achieved by proposing two parallel lightweight branches: the Content-guided Branch (CB) and the Interaction-guided branch. The CB leverages the product object contents to guide the generation of reasonable human-object interaction and ensure consistency between input and output products; The IB, on the other hand, uses both human-pose information and product content information to further enhance the generation of human-object interaction. Equipping with theses two branches, \textit{VirtualModel} supports various product categories(\eg, cosmetics, clothes, shoes, accessories, electronic products, \etc) and types of human-object interactions (\eg, hold, lift, wear, lie, lean on, \etc.). In summary, the main contributions of this paper are as follows:
\vspace{0.5em}
\begin{itemize}
  \item We propose a new task for real-life e-commerce marketing application, named \textit{Object-ID-retentive Human-object Interaction image Generation} (OHG), which focuses on the image quality, reasonability of human-object interaction and another important dimension: the consistency of user-specified product between input and output.
  \item We collect and build a large-scale \textit{HoIHuman} dataset for OHG, which is with high quality/resolution, diverse products/scenarios/human-object-interactions, \etc.
      \vspace{0.5em}
  \item We propose a \textit{VirtualModel} framework with two parallel lightweight branches: content-guided and interaction-guided branches, to explicitly encourage the geneartion of reasonable/high-quality human-object-interaction and to guarantee the ID-consistency of user-specified products.
\end{itemize}
\vspace{0.5em}
Extensive experiments have been conducted on OHG test-set to demonstrate the effectiveness of the proposed \textit{VirtualModel}. Both qualitative visual results and quantitative results on a series of evaluation metrics covering image quality, pose accuracy, text-image alignment, and object-ID consistency have been reported. These show that our \textit{VirtualModel} significantly outperforms other state-of-the-art solutions, such as Controlnet \cite{zhang2023adding}, T2I-adapter \cite{mou2023t2i} and HumanSD \cite{ju2023humansd}, regarding image quality, pose control, reasonable human-object interaction and especially the support for the user-specified product.
\vspace{-1em}
\section{Related Works}
Some related research works are summarized into three parts as follows:

\textbf{Text-to-Image Diffusion Models.} Text-to-image (T2I) generation models, which generate impressive images under the guidance of natural language descriptions, has made remarkable progresses in recent years. Being of the superior scalability and training stability, diffusion-based T2I models have surpassed the conventional GAN-based models in terms of image quality and creativity \cite{dhariwal2021diffusion}. It has become the popular and powerful model in generative family. For example, Stable Diffusion \cite{rombach2022high} and DALL-E2 \cite{ramesh2022hierarchical} show their powerful ability in image generation and become the predominant choice in T2I field. Then DeepFloyd \cite{deepfloyd} and SDXL \cite{podell2023sdxl} further improve the quality and resolution of generated images by several training and model architecture improvements. However, these above models still fails to generate hyper-realistic human images. It might be that the inherent structural information of human are not well captured and learned, such that the generated human parts are unreasonable, \eg, with incorrect number/layout of arms/fingers/legs.

\textbf{Controllable Human Image Generation.} In order to address the above problem, diffusion-based controllable human image generation turns to be focused. For example, ControlNet \cite{zhang2023adding} and T2I-adapter \cite{mou2023t2i} intends to introduce additional trainable modules for receiving pose guidance to the pre-trained text-to-image SD model. Moreover, HumanSD \cite{ju2023humansd} proposes to use heatmap-guided denoising loss to control the learning of human instead of using additional trainable branches. However, all of them still suffer from inaccurate pose control and low-quality human images, and all of them cannot generate human being interacting with specified objects, such that the current human generation technique cannot be applied in actual real-life scenario, \eg, e-commerce marketing. To this end, this paper target at OHG task for e-commerce marketing and intends to generate high-quality human-object interaction images.

\textbf{Datasets for Human Image Generation.} Human-centric datasets like Market1501 \cite{zheng2015scalable}, DeepFashion \cite{liu2016deepfashion} and MSCOCO \cite{lin2014microsoft} have noisy paired source-target images, limited scenarios and low-quality of images. Recently, Human-Art \cite{ju2023human} provides 50K human-centric images in both natural and artificial scenes. However, the number and quality of images fall far short of our needs, \ie, needs for e-commerce marketing where the image should be with high-quality and aesthetic, human parts should be clear and realistic, and the interaction between human and specified-object should be reasonable. Therefore, we collect and build a large scale \textit{HoIHuman} datasets for e-commerce marketing scenario.
\vspace{-1em}
\section{Method}
In this section, we will first describe our defined new task OHG in section \ref{sec_ohg}, then introduce some pre-requisites of diffusion models in section \ref{sec_df}, and finally introduce our proposed \textit{VirtualModel} framework in section \ref{sec_virtualmodel}
\subsection{Task Definition of OHG}\label{sec_ohg}
\begin{figure}
  \centering
  \includegraphics[width=0.8\linewidth]{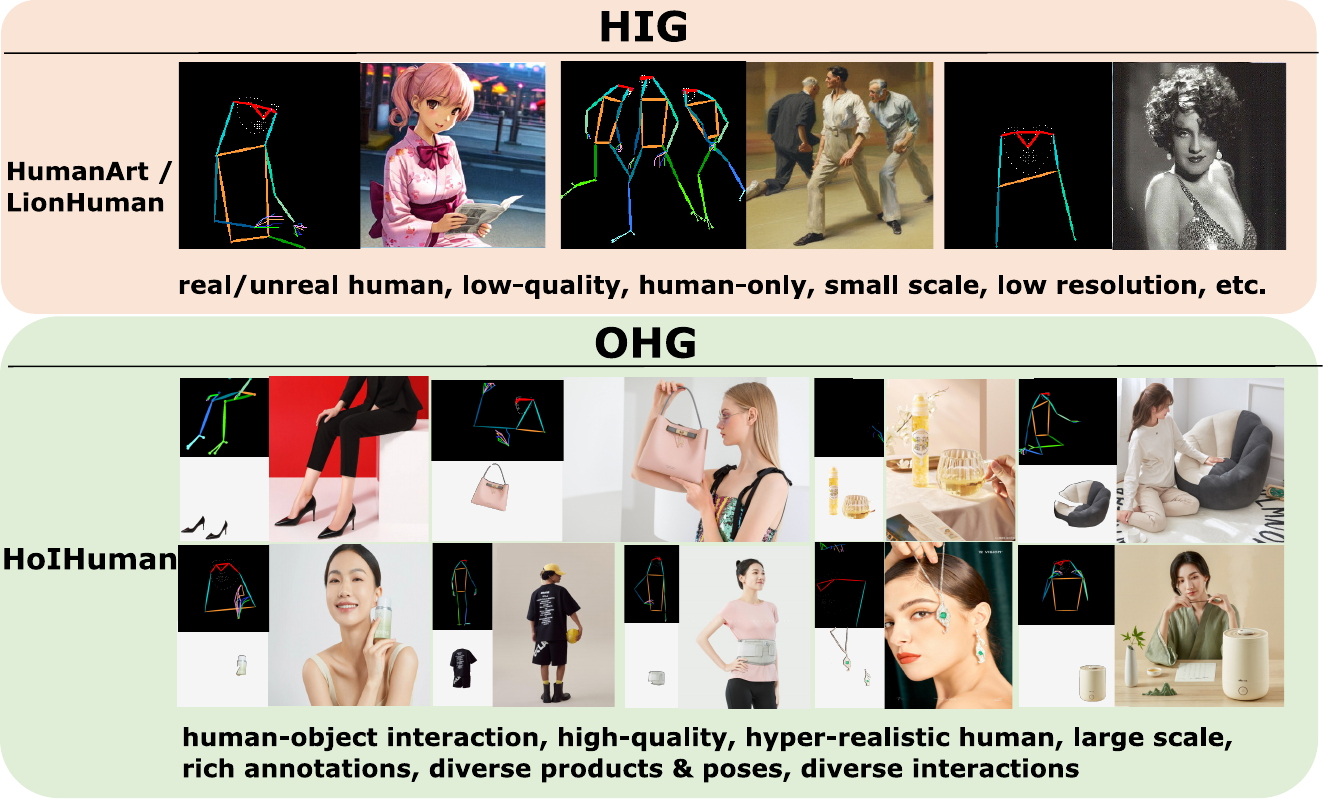}\vspace{-1em}
  \caption{Comparisons of tasks (HIG and our proposed OHG) and the corresponding datasets (HumanArt/Laion-Human\cite{ju2023humansd} and our HoIHuman.)}\label{fig_ohg}
  \vspace{-1em}
\end{figure}
The conventional controllable human image generation (HIG) has been explored many years, in which only human pose-skeleton constraint is provided and the output human is required to be with the same pose as input constraint. However, this task still has gap to some real-life applications, \eg, e-commerce marketing where human always serves as a super-model to display products for sale. In other words, e-commerce scenario requires (1) \emph{generating hyper-realistic human} and (2) \emph{the generated human could interact with the specified products}. To this end, we first define a new human generation task specifically for the e-commerce marketing scenario, called \textit{Object-ID-retentive Human-object Interaction image Generation} (OHG). In which, inputs of this task are the user-specified products and the corresponding human pose information (like pose-skeleton), and outputs are the generated human images containing reasonable interactions with exactly the same product as input. Figure. \ref{fig_ohg} shows the differences between our OHG task and the conventional HIG task, and one can observe that our OHG are built specifically for the e-commerce scene.
\vspace{-1em}
\subsection{Overview of Diffusion Model}\label{sec_df}
\begin{figure*}[!htp]
  \centering
  \vspace{-1em}
  \includegraphics[width=1\linewidth]{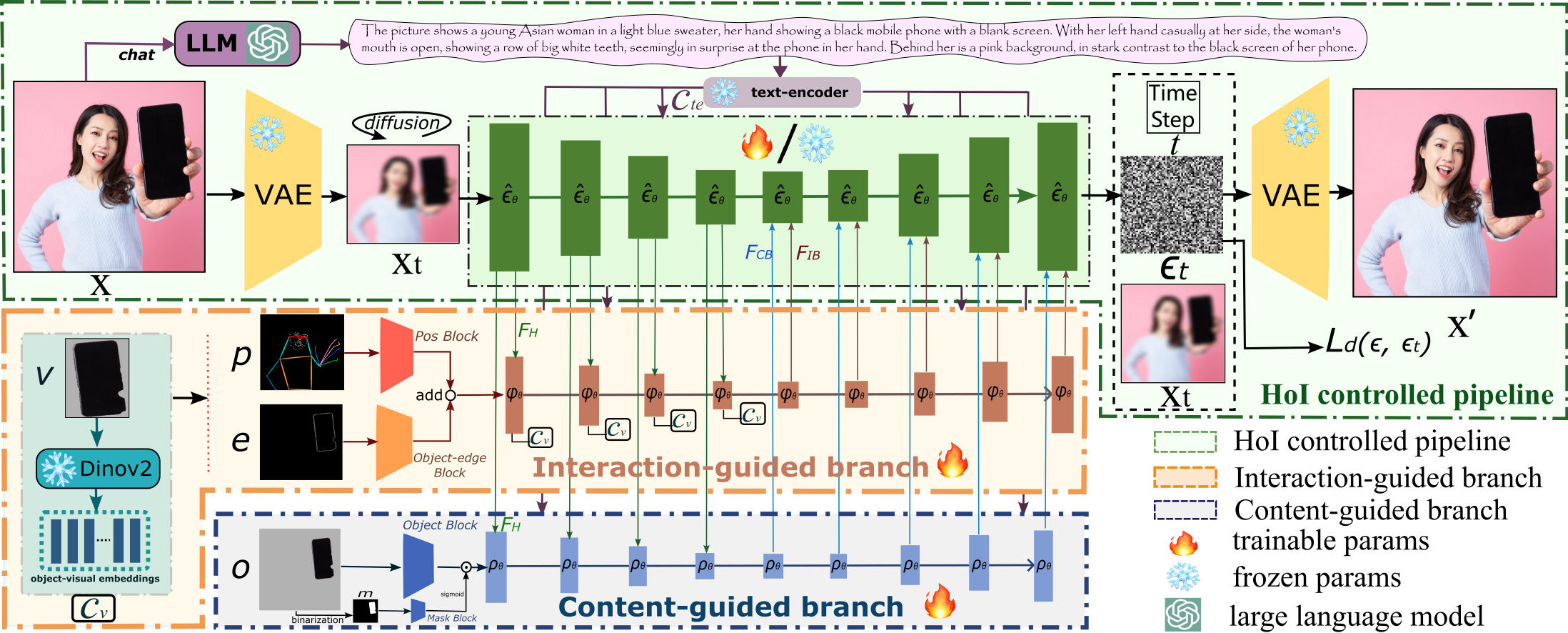}\vspace{-0.5em}
  \caption{Overall framework of the proposed \emph{VirtualModel}, which consists of the Human-object-interaction (HoI) controlled pipeline, the interaction-guided branch and the content-guided branch. During training, paired data are built and fed into \emph{VirtualModel} where $\textsl{x}, o,p,e,v$ are original image and the corresponding image conditions: product \textbf{o}bject, \textbf{p}ose skeleton of human, \textbf{e}dge of product object, close \textbf{v}iew of product, respectively. The text condition $C_{te}$ is obtained by large language model.}\label{fig_vm}
  \vspace{-1em}
\end{figure*}
Diffusion models \cite{ho2020denoising,rombach2022high} are in the field of generative models where the models learn the target
distribution through an iterative denoising process. They comprise two processes: a forward process (also known as the diffusion process) which gradually injects Gaussian noise $\epsilon$ into the raw image data $\textsc{x}$ using a pre-defined Markov chain of T steps, resulting in noised data $\textsc{x}_{T}$, and a learnable denoising process (also known as the reverse process) which converts $\textsc{x}_{T}$ back to $\textsc{x}$ iteratively. Diffusion models can be conditioned on various signals such as class labels, texts or images. Generally, the training objective of a conditional diffusion model $\mathbf{\hat{\epsilon}_{\theta}}(\cdot)$, which predicts noise, is defined as a simplified variant of variational bound:
\begin{equation}\label{eq_ddpm}
  L_{d} = {\mathbb{E}_{\mathbf{\textsc{x}},\mathbf{c},\mathbf{\epsilon},t}[w_{t}\|\mathbf{\hat{\epsilon}}_{\mathbf{\theta}}(\alpha_{t}\textsl{x}+\sigma_{t}\mathbf{\epsilon};\mathbf{c})-\mathbf{\epsilon}\|^{2}_{2}]}
\end{equation}
where $\mathbf{\textsl{x}},\mathbf{c}$ are the sample/condition pairs sampled from the training distribution; $\mathbf{\epsilon}\sim\mathcal{N}(\mathbf{0},\mathbf{I})$ is the ground-truth gaussian noise. $t\sim \mathcal{U}[1,T]$ is training time-step at each iteration; $\alpha_{t},\sigma_{t},w_{t}$ are hyper-parameter terms that control the noise-adding schedule and sample quality decided by the diffusion sampler. At training stage, $\mathbf{\hat{\epsilon}_{\theta}}(\cdot)$ is optimized to predict the noise $\mathbf{\epsilon}$ that corrupts $\textsl{x}$ into $\textsl{x}_{t}$: $\textsl{x}_{t}=\alpha_{t}\textsl{x}+\sigma_{t}\mathbf{\epsilon}$. At inference stage, data samples can be generated from Gaussian noise $\textsl{x}_{T}\sim\mathcal{N}(\mathbf{0},\mathbf{I})$ using DDPM \cite{ho2020denoising} or DDIM \cite{song2020denoising}.
\vspace{-1em}
\subsection{VirtualModel}\label{sec_virtualmodel}
In order to generate hyper-realistic product advertising images shown by human-model, we propose \emph{VirtualModel} for OHG. Figure \ref{fig_vm} depicts the overall architecture of \emph{VirtualModel}, which contains Human-object-interaction (HoI) controlled pipeline and two primary modules (the interaction-guided branch and the content-guided branch). Our method is trained by paired data where $\textsl{x}, o,p,e,v$ are original image and the corresponding conditions: product \textbf{o}bject, \textbf{p}ose skeleton of human, \textbf{e}dge of product object, close \textbf{v}iew of product, respectively. During inference time, $o,p$ are first given to our method, then conditions $e,v$ will be produced by edge detector and simple image-crop, respectively.
\vspace{-1em}
\subsubsection{HoI Controlled Pipeline:}
Following the successful Stable Diffusion \cite{rombach2022high} method, we obtain the latent representation $\textsl{x}_{0}\in\mathbb{R}^{h\times w\times c}$ of original input image $\textsl{x}$ via VAE encoder. Then, in the diffusion process, Gaussian noise is added to obtain $\textsl{x}_{t}=\alpha_{t}\textsl{x}_{0}+\sigma_{t}\mathbf{\epsilon}$ at time step $t$. To learn a human-object interaction controllable model, parameterized by Unet $\hat{\epsilon}_{\theta}$, we employ several conditions: text embeddings $C_{te}$ which are obtained by chatting with the large language model QwenVL \cite{qianwenVL} (ask it by `please describe this image in detail, including human, object and background') and then encoding prompts with CLIP, interaction feature $F_{IB}$ from interaction-guided branch, and object feature $F_{CB}$ from content-guided branch. Then our \emph{VirtualModel} can be optimized with Equation \ref{eq_ddpm} by replacing $\mathbf{c}$ with ($C_{te},F_{IB},F_{CB}$). Specifically, text conditions are employed in each cross-attention layers in the Unet, and conditions ($F_{IB}, F_{CB}$) are added on the input to each transformer block in output$\_$blocks\footnote{Named in pytorch format} of Unet.

While, considering that our OHG task requires (1) the generated human should be hyper-realistic (2) the interaction between human and product object should be reasonable and consistent with the conditions. We experimentally found and employed a decoupled training strategy: training base Unet $\hat{\epsilon}_{\theta}$ and other control branches separately. It is because that the base Unet mainly takes charge of the quality/beauty of contents, while other branches mainly focus on the precision of guidance. Therefore, we can use different specific data to optimize different components to be the best. Specifically, we first train the base Unet by using close view image crops of human to enhance the generated human quality. Then freeze the base Unet and train CB and IB by using diverse views of images containing both human and product object. We experimentally found using this strategy can guarantee both the generation quality of human parts (\eg, face/hand/arm/leg) and the reasonability of interaction between human and object.
\vspace{-1em}
\subsubsection{Interaction-guided Branch:}
As shown in Figure \ref{fig_vm}, the Interaction-guided Branch (IB) has three inputs, including pose skeleton image $p$, object edge image $e$ and the close view of product image $v$. Specifically, pose $p$ and edge $e$ images are fed into the corresponding separate expert blocks (seven 3x3 convolution layers with channels [16,16,32,32,96,96,256]) to extract the specific low-level features. These features are of precise controllable raw information and are fused together by addition operation, and then are fed into a lightweight Unet $\phi_{\theta}$ (same architecture as $\hat{\epsilon}_{\theta}$ yet with fewer channels). To capture more raw information from original image $\textsl{x}$, $\phi_{\theta}$ adds the output features $F_{H}$ from each input$\_$blocks of the base Unet on its own corresponding block's input features. And to precisely control the interaction between human and object, each output$\_$blocks' features of $\phi_{\theta}$ are finally added to the base Unet. It is worth noting that $p$ and $e$ can guide Unet to model reasonable human posture and object position, respectively. And by merging them together, the interaction information between human and object can be easily obtained by deep models. However, edges of different objects might be confused especially when some products have similar shapes, simply using $p,e$ conditions might be not enough to model reasonable HoI. Therefore, we introduce condition $v$ to further enhance the perception of objects. As in Figure \ref{fig_vm}, $v$ is fed into DinoV2 \cite{oquab2023dinov2} and the penultimate features are extracted as $C_{v}$. $C_{v}$ is fed into a linear transformation layer $\xi$, and then is used the same as text condition $C_{te}$ but in another parallel cross-attention layer as:
\begin{equation}
  f=CrossAtten^{1}(f,C_{te})+CrossAtten^{2}(f,\xi(C_{v}))
\end{equation}
where $f$ is the propagated image feature from the previous layer, $CrossAtten^{i},~i\in[1,2]$ are conventional cross-attention layers. We experimentally found using $C_{v}$ in input$\_$blocks of $\phi_{\theta}$ is enough. Since if also injecting $C_{v}$ in output$\_$blocks of $\phi_{\theta}$, the extra computation costs are doubled but with few performance improvements. And in order to reduce the GPU memory cost and to speed up the training of IB and CB branches, the visual embeddings $C_{v}$ for each data pairs will be extracted offline ahead of time.
\vspace{-1em}
\subsubsection{Content-guided Branch:}
IB branch focuses on the perception of HoI, yet still is weak on maintaining the product-ID consistency. While in e-commerce marketing scenario, the product shown in the advertising images should be exactly same with the product itself provided by seller. To this end, we introduce a Content-guided Branch (CB) to explicitly impose product content constraint into the diffusion model. Specifically, condition $o$ is first processed by binarization, producing mask image $m$. $o,m$ will be fed into the 7-layers' object-block and 3-layers' mask-block\footnote{Channels are [16,16,32,32,96,96,256] and [32,32,1], respectively}, respectively. And their output features will be merged by Hadamard Product $\bigodot$. These two blocks are employed to extract the precise low-level visual information about the provided product object. Then, as in Figure \ref{fig_vm}, we also employ a lightweight Unet with the same model architecture as in IB, parameterized by $\rho_{\theta}$, to perform the product-content guidance. And the information flow is propagated the same as in IB.
\vspace{-1em}
\section{HoIHuman Dataset}
\begin{figure}[t]
\vspace{-1em}
  \centering
  \includegraphics[width=0.5\linewidth]{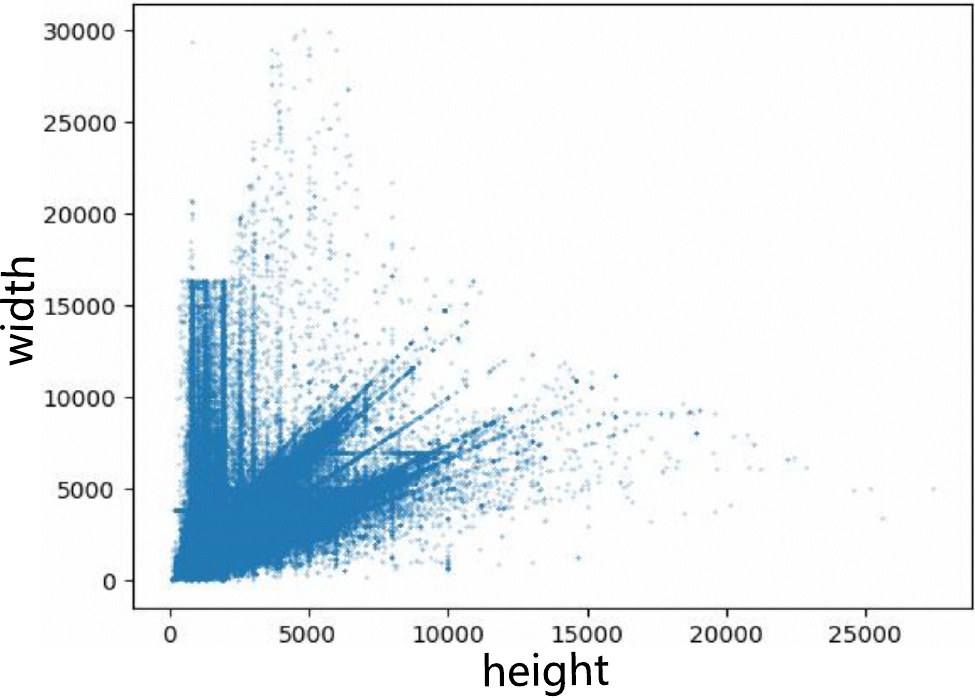}
  \vspace{-1em}
  \caption{Resolution statistics of our HoIHuman dataset. One can observe that our HoIHuman is with high resolution and high quality images.}\label{fig_hoihuman}
  \vspace{-1em}
\end{figure}

As the conventional human generation datasets, like Market1501 \cite{zheng2015scalable}, DeepFashion \cite{liu2016deepfashion}, MSCOCO \cite{lin2014microsoft}, HumanArt and LaionHuman \cite{ju2023human,ju2023humansd} are of low-quality, human-only, small scale and low resolution, and more serious is that the human might be unreal human such as cartoon human, figure sculpture and so an. Therefore, to train a diffusion model for generating hyper-realistic human-object interacted images especially for e-commerce scenario, we collect and build a large-scale dataset, called HoIHuman. Specifically, we collect specific e-commerce images internally. The images are then filtered by YOLOX person detection \cite{ge2021yolox}, resolution (with shortest side $>$ 256) and products (cosmetics, clothes, shoes, accessories, electronic products, etc). We employ OCR \cite{ocr} and Lama \cite{suvorov2021resolution,lama} to remove the watermark/logo/text on the images. Then each image is given by a text description by QWenVL \cite{qianwenVL}. We use ViTPose \cite{xu2022vitpose}, Grounding-DINO \cite{liu2023grounding} and SAM \cite{kirillov2023segment} to automatically obtain the pose-skeleton and product-object region annotations. After doing all the above, we obtain 3,156,125 images with rich annotations and high quality, and the resolution distribution is shown in Figure \ref{fig_hoihuman}. Moreover, we select 5k images out for testing. More details can be found in supplementary materials.
\vspace{-1em}
\section{Experiments}
\textbf{Implementation Details:} We follow classifier-free guidance \cite{ho2022classifier} and train our models with conditioning dropout: each conditional inputs are dropped out with 0.1 probability. The batch size for training base Unet and other branches are 256 and 128, respectively. Adam optimizer \cite{kingma2014adam} is employed and learning rate is set to 1e-5. All experiments are conducted on NVIDIA-32G-V100 GPUs. During inference, DDIM sampler \cite{song2020denoising} with 30 steps is adopted and the CFG is set to 7.0 default. In order to save GPU memory costs during training, we can pre-extract the latent image features $\textsl{x}_{0}$ offline as well as the text embedding $C_{te}$. During inference, to make sure the generated object to be exactly the same with input object, we crop the original object based on the mask $m$ and then copy-paste it to the position of generated object. This kind of post-processing can make sure the contents within mask $m$ to be exactly the same with original input, and we call it \emph{\textbf{Content Backfill}} (CBF) post-processing strategy.

\textbf{Evaluation Metrics:} In order to illustrate the effectiveness of our proposed \emph{VirtualModel}, we use several metrics covering: (1) \textbf{image quality} which is evaluated by the widely used metrics of Fr$\acute{e}$chet Inception Distance \cite{heusel2017gans} (FID) and Kernel Inception Distance \cite{binkowski2018demystifying} (KID), (2) \textbf{pose accuracy} which is evaluated by distance-based Average Precision (AP) and Average Recall (AR), (3) \textbf{text-image consistency} which is evaluated by the CLIP similarity between text and image embeddings, and (4) \textbf{product-ID consistency} which is evaluated by Object Extension Ratio (OER) metric that proposed by us especially for OHG task.
\begin{figure}[t]
  \centering
  \includegraphics[width=0.9\linewidth]{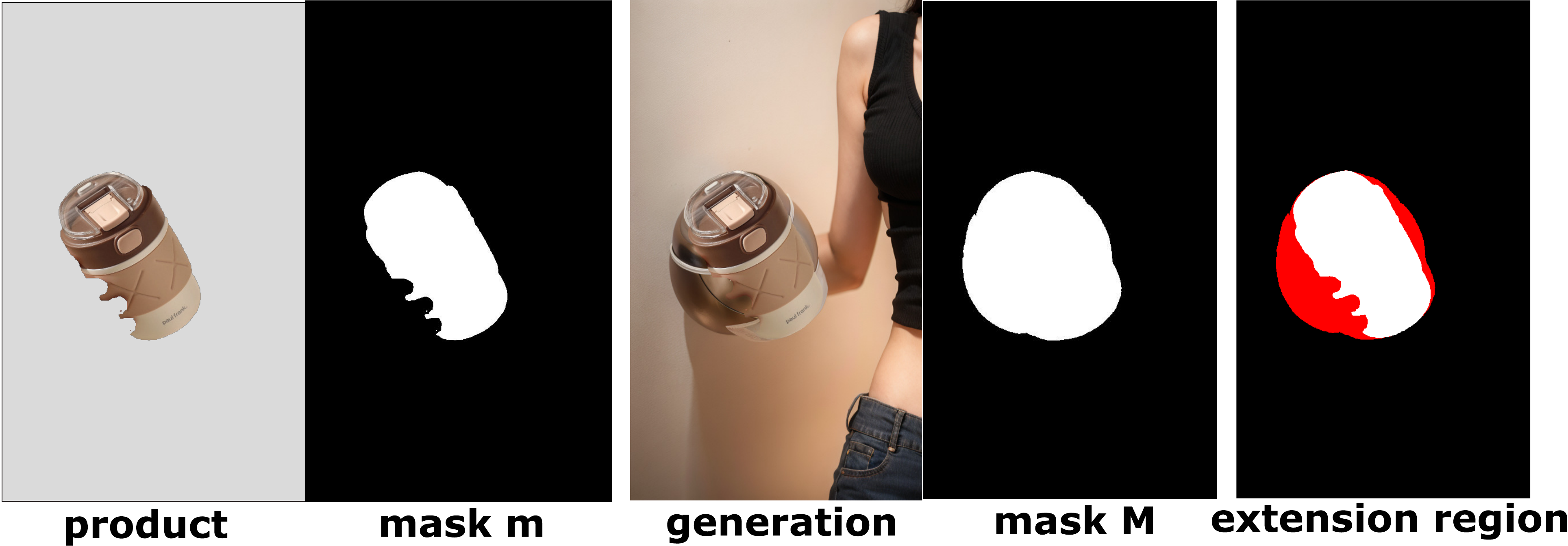}\vspace{-1em}
  \caption{Illustration of the proposed metric Object Extension Ratio (OER) computation. Red region is the extended content which is incorrect.}\label{fig_oer}\vspace{-1em}
\end{figure}

Since we can adopt CBF post-processing, the consistency of content within mask $m$ can be guaranteed. But the consistency of content without mask region $m$ cannot be guaranteed. To this end, as shown in Figure \ref{fig_oer}, we perform SAM segmentation on the generated images (after CBF) to obtain a new mask $M$ for the specified product, so as to detect whether there are new contents generated based on the given product. And the Object-Extend-Ratio (OER) can be computed as:
\begin{equation}\label{eq_oer}
  OER=\frac{\sum ReLU(M-m)}{\sum m}
\end{equation}
where $M,m$ is 0-1 binary masks, ReLU is the activation function. From equation \ref{eq_oer}, one can observe that OER is ranged from 0 to $+\infty$, and the smaller OER is, the better performance the method has.

\textbf{Comparison methods:} Despite this paper focuses on a novel OHG task, we try our best to compare with other methods: including the Text-to-Image methods like Stable Diffusion 1.5 \cite{rombach2022high}, Stable Diffusion 2.1 \cite{rombach2022high}, SDXL \cite{podell2023sdxl} and DeepFloyd IF \cite{deepfloyd}, and the pose controllable human image generation methods like ControlNet \cite{zhang2023adding}, T2I-adapter \cite{mou2023t2i} and HumanSD \cite{ju2023humansd}. Moreover, we also build a MultiControls ControlNet with pose and inpainting conditions so as to implement the OHG task. It is worth noting that, for fair comparison, all the inference hyper-paramters used in these methods like CFG, DDIM steps, output image resolution, negative prompts are the same as our \emph{VirtualModel}.
\vspace{-1em}
\subsection{Quantitative Analysis}
\begin{table*}[htbp]
  \centering
  \vspace{-1em}
  \caption{Quantitative comparisons between our \emph{VirtualModel} and other diffusion models (including text-to-image (T2I) and condition-to-image (C2I) methods) over HoIHuman-5k test set. All the other methods are evaluated using their officially released models and codes. The CLIP score is computed by openai/clip-vit-base-patch16 model. It is worth noting that AP and AR metrics for pose evaluation are computed on the challenging 133 whole body key points instead of the 17 coarse key points used in other papers.}\vspace{-1em}
  \resizebox{1.0\linewidth}{!}{
    \begin{tabular}{c|l|c|cccccc}
    \hline
    \multicolumn{2}{c}{\multirow{2}[2]{*}{Methods}} & \multirow{2}[2]{*}{Task} & \multicolumn{2}{c}{Image Quality} & Text-Image Consistency & \multicolumn{2}{c}{Pose Accuracy} & Object-ID Consistency \\
\cmidrule{4-9}    \multicolumn{2}{c}{} &       & FID $\downarrow$   & KID$_{\times 1k}$ $\downarrow$ & CLIP $\uparrow$  & AP $\uparrow$   & AR $\uparrow$   & OER($\%$) $\downarrow$ \\
\hline
    \multirow{4}[0]{*}{T2I} & SD15 \cite{rombach2022high}  & HIG   & 30.31 & 9.97  & 23.34 & -    & -    & - \\
          & SD2.1 \cite{rombach2022high} & HIG   & 28.26 & 10.02 & 23.27 & -    & -    & - \\
          & SDXL \cite{podell2023sdxl} & HIG   & 25.22 & 6.66  & \textbf{24.97} & -    & -    & - \\
          & DeepFloyd IF \cite{deepfloyd}& HIG   & 29.89 & 11.75 & 24.51 & -    & -    & - \\
          \hline
    \multirow{5}[0]{*}{C2I} & ControlNet \cite{zhang2023adding} & HIG   & 25.87 & 8.00     & 23.13 & 3.89  & 13.05 & - \\
          & T2I-Adapter \cite{mou2023t2i}& HIG   & 25.79 & 7.75  & 24.48 & 3.40   & 11.79 & - \\
          & HumanSD \cite{ju2023humansd}& HIG   & 29.22 & 9.31  & 24.01 & 10.38 & 23.16 & - \\
          \cmidrule{2-9}
          & MultiControlNet \cite{zhang2023adding}& OHG   & 24.62 & 7.84  & 23.65 & 5.34  & 14.71 & 32.73 \\
          & \textbf{VirtualModel (ours)} & OHG   & \textbf{18.37} & \textbf{5.92}  & 24.17 & \textbf{31.74} & \textbf{51.28} & \textbf{1.71} \\
          \hline
    \end{tabular}%
    }\vspace{-1em}
  \label{tab:sota}%
\end{table*}%
\begin{table*}[!tbp]
  \centering
  \caption{Quantitative Results on Human Preference-Related Metrics for image quality evaluation. Related metric PickScore is reported. It means the ratio of preferring ours to others, where number larger than 50$\%$ means our method is better and the larger the number, the better our method is than the corresponding method. One can observe that our \emph{VirtualModel} achieves the best performance.}
  \resizebox{1.0\linewidth}{!}{
    \begin{tabular}{lcccccccc}
    \hline
    Methods & \multicolumn{1}{l}{Ours to:} & \multicolumn{1}{l}{SD1.5\cite{rombach2022high}} & \multicolumn{1}{l}{SDXL\cite{podell2023sdxl}} & \multicolumn{1}{l}{DeepFloyd IF\cite{deepfloyd}} & \multicolumn{1}{l}{ControlNet\cite{zhang2023adding}} & \multicolumn{1}{l}{T2I-adapter\cite{mou2023t2i}} & \multicolumn{1}{l}{HumanSD\cite{ju2023humansd}} & \multicolumn{1}{l}{MultiControlNet\cite{zhang2023adding}} \\
    \hline
    PickScore &       & 88.97$\%$ & 76.04$\%$ & 77.66$\%$ & 73.13$\%$ & 72.2$\%$  & 75.86$\%$ & 68.24$\%$ \\
    \hline
    \end{tabular}%
    }
    \vspace{-1em}
  \label{tab_score}%
\end{table*}%
Firstly, we conduct quantitative evaluations on HoIHuman-5k test set. The evaluation results are shown in Table \ref{tab:sota}. For all methods, we use the default CFG scale of 7.0, which well balances the quality and diversity with appealing results. And DDIM sampling steps are set to 30. For T2I methods, we set the output image resolution to $512\times 512$ (if the output resolution is $1024\times 1024$ like in SDXL \cite{podell2023sdxl} and DeepFloyd IF \cite{deepfloyd}, we will resize images to $512\times 512$). For C2I methods, the output images are with the same aspect-ratios as input images, and the shortest side is set to 512. From Table \ref{tab:sota}, one can observe that our \emph{VirtualModel} outperforms all the other competing methods in terms of image quality, pose accuracy and object-ID consistency by a large margin, and achieves on-par text-image consistency score. Note that SDXL \cite{podell2023sdxl} and DeepFloyd IF \cite{deepfloyd} use more powerful text encoders and more and larger Unets, thus leading to superior text-image consistency. In spite of this, we still obtain on-par CLIP score.
Moreover, we use another human preference-related metrics to demonstrate \emph{VirtualModel} performances: \ie, PickScore \cite{kirstain2023pick}, which is trained on the side-by-side comparisons of two T2I models. The results are reported in Table \ref{tab_score}. \footnote{Note that, to prevent the evaluation models from overfitting on its own training data, we add some high quality images to finetune the used evaluation models.}One can observe that our \textsl{VirtualModel} achieves the best preference performances.
\begin{table}[!tbp]
  \centering
  \caption{User-Study on HoIHuman 1k-test set. The volunteer is given four images simultaneously, and order of these images are shuffled each time for fair comparison. The volunteer will choose the best one of the four. Then results from 3 volunteers are averaged. Num refers to the average number of wins.}
    \begin{tabular}{ccccc}
    \hline
          & \multicolumn{1}{l}{T2I-adapter\cite{mou2023t2i}} & \multicolumn{1}{l}{ControlNet\cite{zhang2023adding}} & \multicolumn{1}{l}{HumanSD\cite{ju2023humansd}} & \multicolumn{1}{l}{Ours} \\
          \hline
    Num   & 37.33    & 43.67    & 1.67     & \textbf{917.33} \\
    \hline
    \end{tabular}%
    \vspace{-0.5em}
  \label{tab_userstudy}%
\end{table}%

However, considering that all the above image quality metrics like FID, KID, text-image alignment CLIP score and PickScore diverge a lot from the actual human preference. To this end, we report user-study on HoIHuman 1k-test set (it is randomly select from the original 5k test set) in Table \ref{tab_userstudy}. From this table, one can observe that our \emph{VirtualModel} significantly outperforms other methods by a large margin.
\vspace{-1em}
\subsection{Qualitative Analysis}
\begin{figure*}[t]
  \centering
  \includegraphics[width=1.0\linewidth]{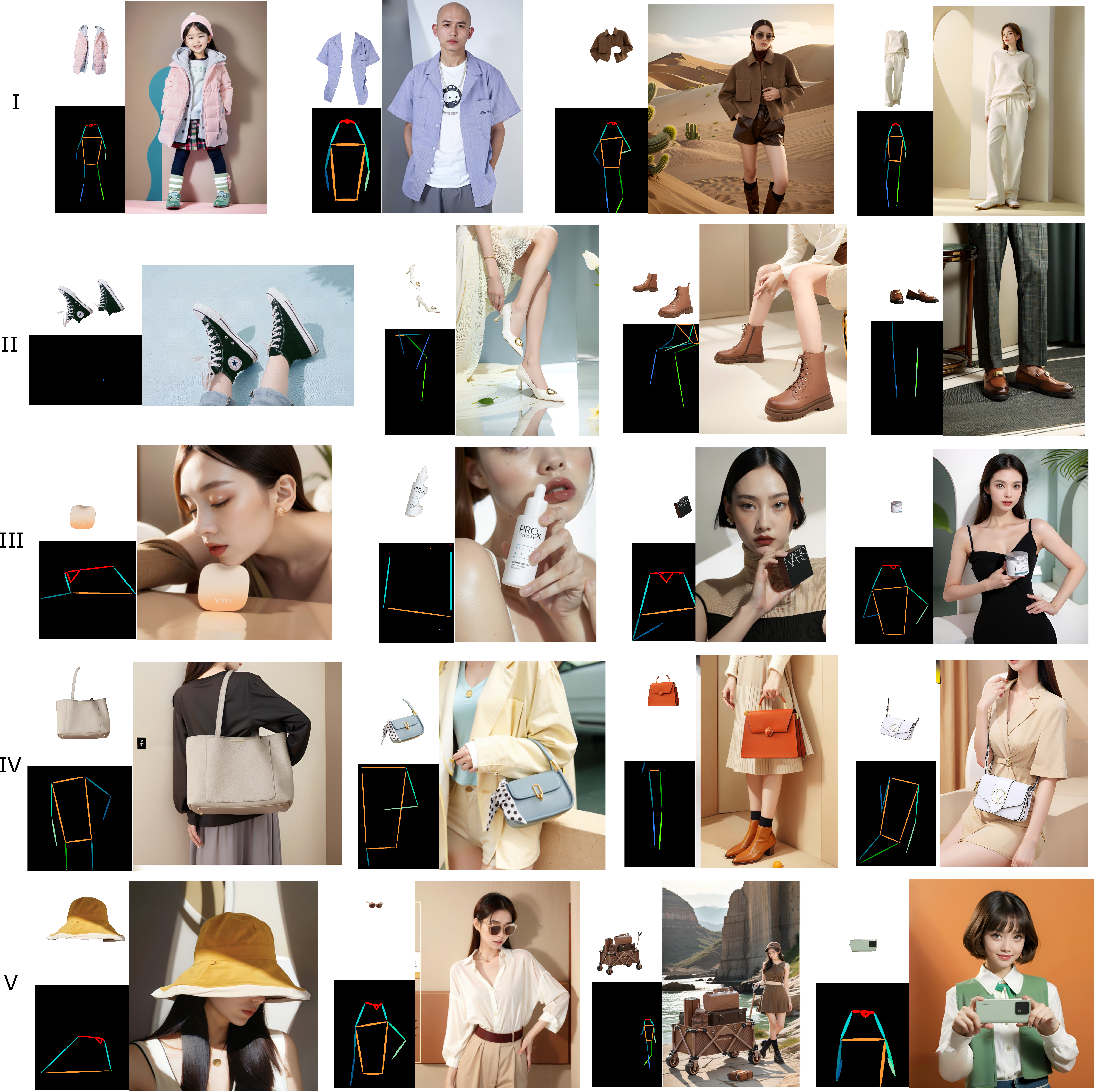}
  \vspace{-0.5em}
  \caption{Qualitative results of \emph{VirtualModel} for OHG task. The products contain (I) clothes, (II) shoes, (III) cosmetics, (IV) bags and (V) so on.}\label{fig_quality}
  \vspace{-2em}
\end{figure*}

Figure \ref{fig_intro} has shown example comparisons with other SOTA methods, from this figure one can observe that other T2I/C2I methods are inferior in terms of pose control, image quality and even the support for product display and ID-consistency maintaining. In addition, to further demonstrate the effectiveness of \emph{VirtualModel}, we show more visual cases (with different category types of products, different interactions between human and object) as in Figure \ref{fig_quality}. From this figure, one can observe that our \emph{VirtualModel} is good at generating hyper-realistic human images for product shown especially in e-commerce scenario.
\subsection{Ablation Study}
\textbf{Effects of IB and CB.} Table \ref{tab_ab} shows the ablation study on IB and CB. One can observe that CB mainly focuses on the constraint of object-ID consistency. Since CB spatially extract the overall information of the provided object and then explicitly control the corresponding spatial position to be the same as this object. Moreover, IB mainly focuses on the constraint of interaction and human pose accuracy. It is because (1) the precise pose annotations are given, such that the overall coarse human pose can be recognized by diffusion model; and (2) object-edge and object-embedding are given to help diffusion model to know which interaction between human and object is reasonable.

\begin{figure}[!tp]
  \centering
  \includegraphics[width=0.65\linewidth]{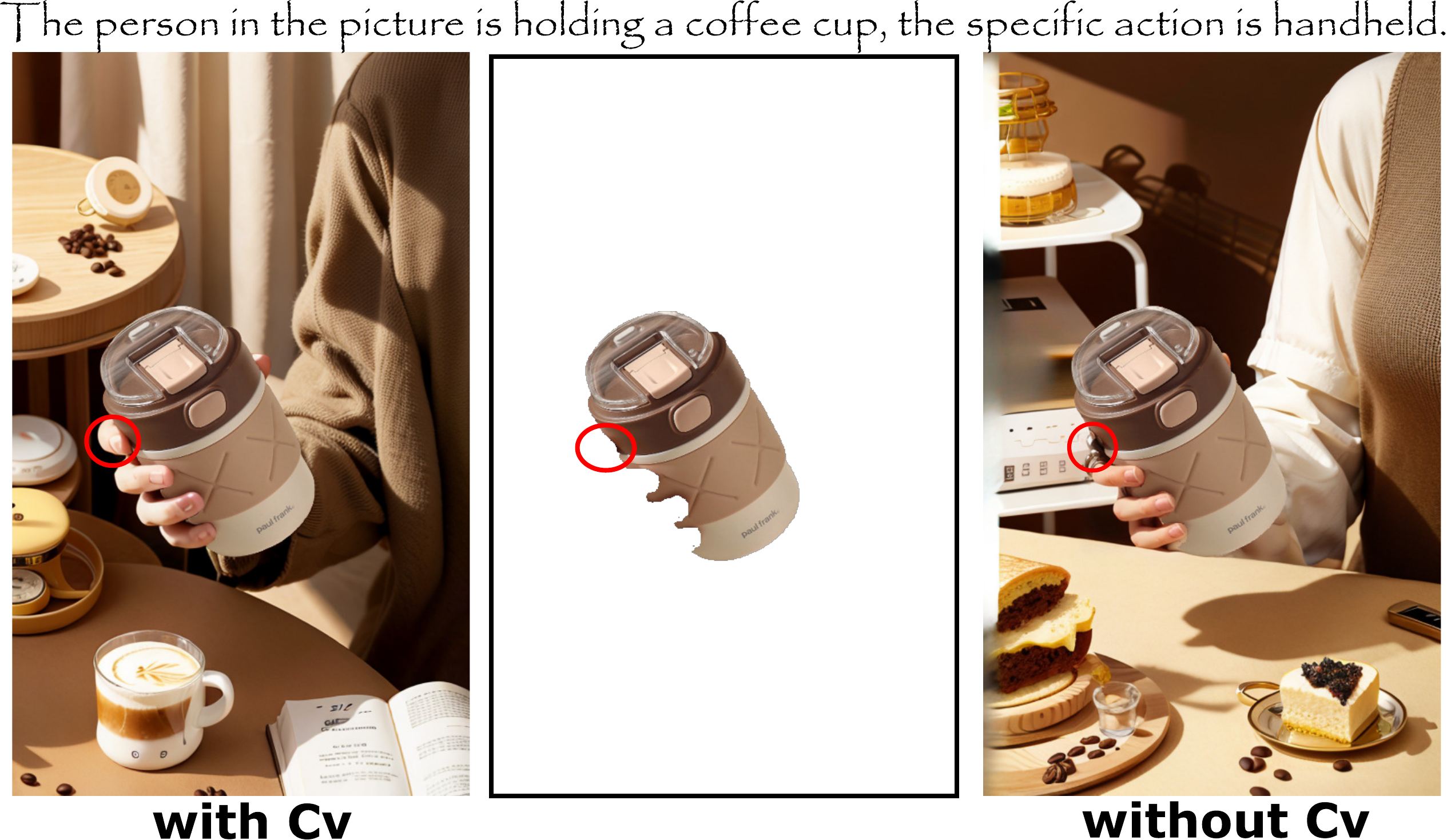}\vspace{-1em}
  \caption{Effect of $C_{v}$. Middle is the user-specified product.}\label{fig_ab_cv}\vspace{-2em}
\end{figure}

\textbf{Effect of $C_{v}$.} Figure \ref{fig_ab_cv} shows the differences with/without $C_{v}$ embedding. Since DinoV2 is a stronger image encoder. It can provide useful features of object to IB to recognize which interaction is reasonable and to help the corresponding content generation.

More ablation studies can be found in supplementary materials.
\begin{table}[tp]
  \centering
  \caption{Ablation study on IB and CB modules.}\vspace{-1em}
    \begin{tabular}{c|ccc}
    \hline
    \multirow{2}[3]{*}{Methods} & \multicolumn{2}{c}{Pose Accuracy} & Object-ID Consistency \\
\cmidrule{2-4}          & AP$\uparrow$    & AR$\uparrow$    & OER($\%$)$\downarrow$ \\
    \hline
    \emph{VirtualModel} & \textbf{31.74} & \textbf{51.28} & \textbf{1.71} \\
    \hline
    w/o CB & 28.66 & 50.77 & 10.25 \\
    w/o IB & 0.7  & 3.6     & 2.93 \\
    \hline
    \end{tabular}%
  \label{tab_ab}%
  \vspace{-1em}
\end{table}%
\vspace{-1em}
\section{Discussion}
\textbf{Conclusion.} In this paper, we propose a novel OHG task for scenario of e-commerce marketing and build a corresponding large-scale dataset. Based on these, VirtualModel is proposed by introducing Interaction-guided Branch and Content-guided Branch. Extensive experiments and visualization cases demonstrate our framework is good at OHG task, obtaining hyper-realistic product shown images by human-model.

\textbf{Limitation and Future Work.} Due to the limited performance of existing pose/detection/segmentation estimators for real word, we find it sometimes fails to generate good results like finger/toe/face, especially when these parts take up only a small part of the overall picture. The current framework still requires pose skeleton as input, and we hope it can be produced by the generation model as well. Thus when product is given, we can first get a reasonable pose and then use it to generate the corresponding product shown images.

\section{Supplymentary}

\begin{figure*}[hbp!]
  \centering
  \includegraphics[width=1.0\linewidth]{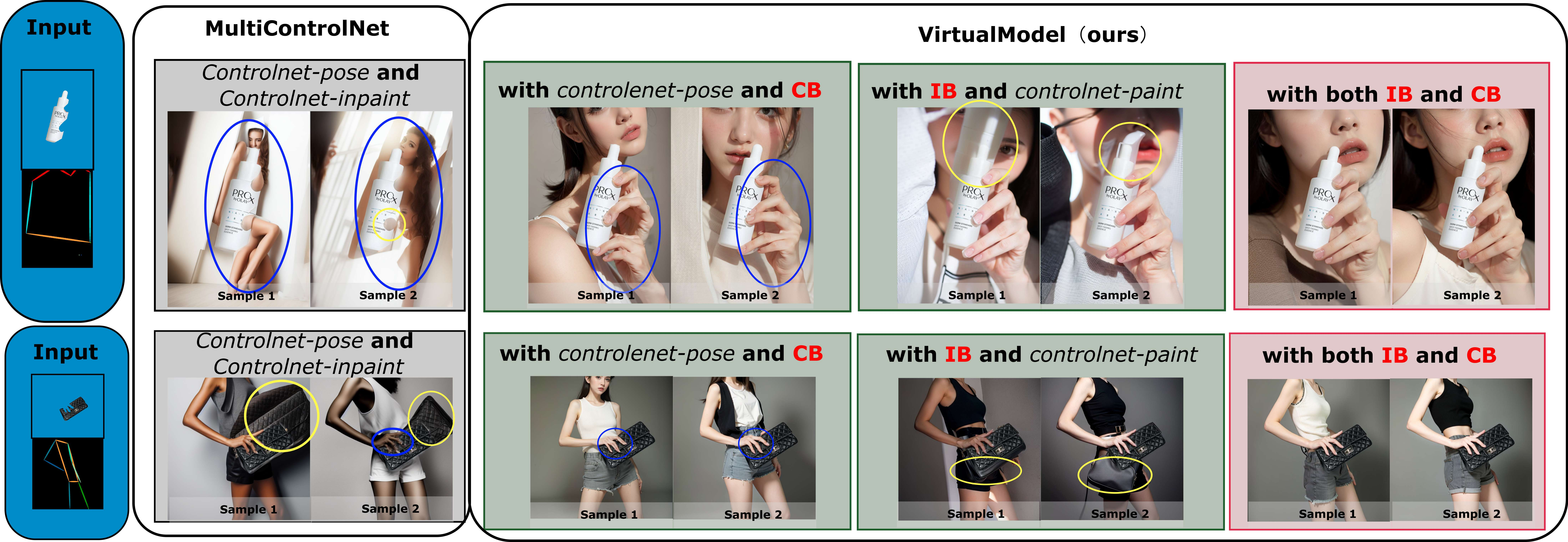}
  		\captionof{figure}{\label{fig_ab_vm} Visual comparisons between our \textit{VirtualModel} and MultiControlNet. The inputs for MultiControlNet and our \textit{VirtualModel} are the same, \ie, masked product objects and pose skeletons. We implement the MultiControlNet using both the ControlNet-Inpaint \cite{zhang2023adding} and ControlNet-Pose \cite{zhang2023adding} models. Furthermore, to demonstrate the effectiveness of our methods, we replace CB and IB with ControlNet-Pose and ControlNet-Inpaint methods, respectively. All the generated images in each row are from the same prompt. \color{blue}{Blue circle} \color{black}{and}  \color{yellow}{yellow circle} \color{black}indicate unreasonable interactions and changes of product identity, respectively. \textbf{Best viewed with zoom-in}.}\vspace{-1em}
\end{figure*}
\begin{figure*}[ht!]
  \centering
  \includegraphics[width=0.95\linewidth]{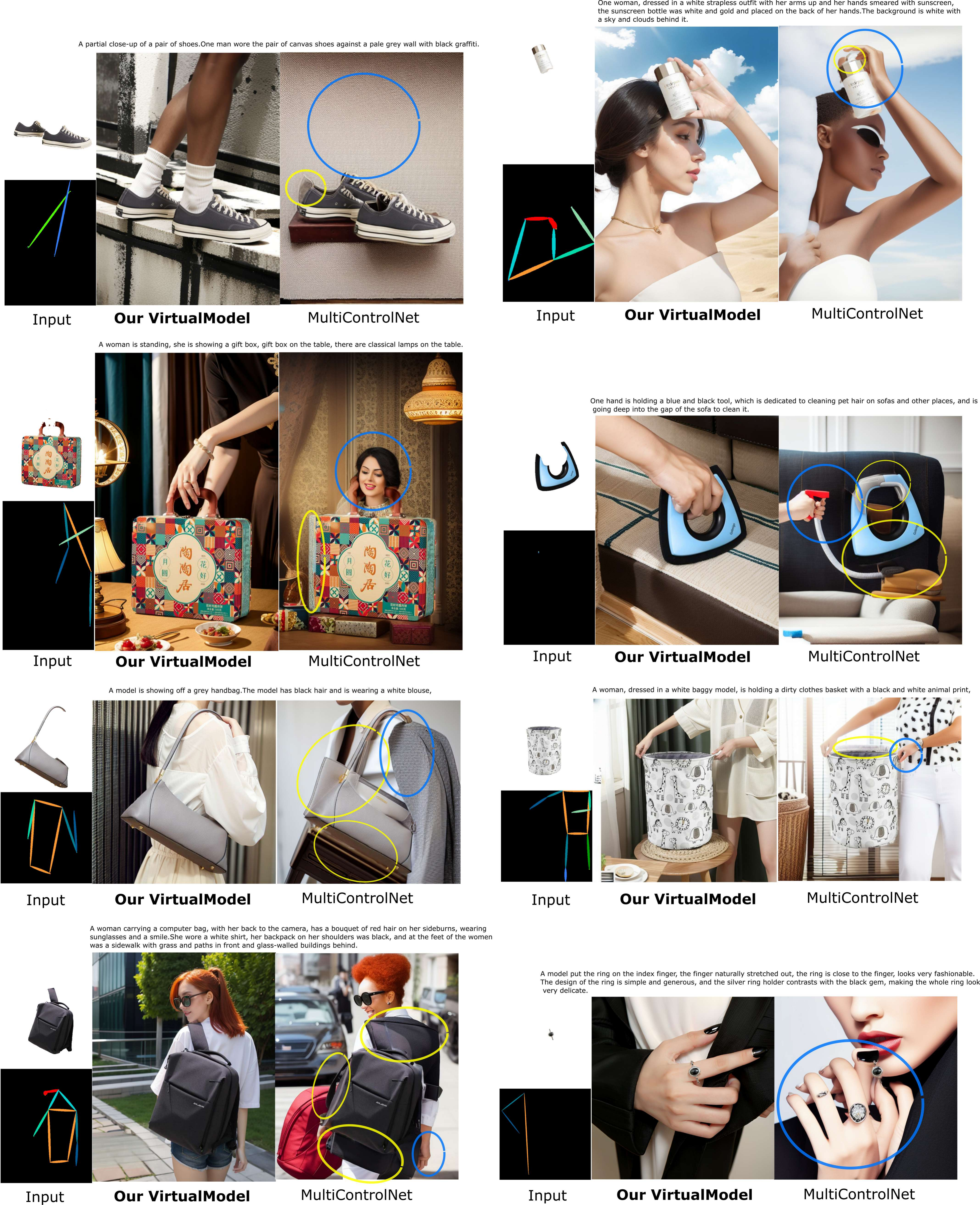}
  \caption{Additional comparison examples between our \emph{VirtualModel} and MultiControlNet. \color{blue}{Blue circle} \color{black}{and}  \color{yellow}{yellow circle} \color{black}indicate unreasonable interactions and changes of product identity, respectively. \textbf{Best viewed with zoom-in}.}\label{fig_morecomp}
\end{figure*}
\section{More Ablation Studies}
\textbf{1. Comparisons between our VirtualModel and MultiControlNet}: In the conventional HIG task, only the pose similarity between the input pose-skeleton condition and the output human image is taken into account. However, this paper addresses a novel OHG task that differs from the conventional HIG task. In other words, it not only considers the pose similarity between the input condition and the output human image, but also ensures that the interactions between the human and the given products are reasonable and realistic, while maintaining the ID-consistency of the given products. As a result, the currently existing methods like ControlNet \cite{zhang2023adding}, T2I-adapter \cite{mou2023t2i} and HumanSD \cite{ju2023humansd} cannot be employed in our OHG task.

To this end, we implement a MultiControlNet framework, where ControlNet-Pose and ControlNet-Inpaint models are jointly used (they focus on human-pose control and content inpainting, respectively), to implement this OHG task. \textbf{Table.1} (in main paper) shows the quantitative comparisons between our \emph{VirtualModel} and this MultiControlNet, one can observe that our method outperforms MultiControlNet by a large margin in all evaluation metrics.

Moreover, to visually observe the differences, we provide comparison examples in Figure \ref{fig_ab_vm}. From the second column images of this figure, one can observe that when using MultiControlNet unreasonable human-object interacting images are generated (indicated by blue circle) and the product identity is changed (indicated by yellow circle). It is because that ControlNet-pose simply considers the control on the coarse human-pose instead of the possible interaction status between human and object, and ControlNet-inpaint also ignores the precise constraint on the generated object contents. Furthermore, in order to highlight the importance of each module in \emph{VirtualModel}, we adopt the principle of controlling variables by using ControlNet-Pose and ControlNet-Inpaint to replace our IB and CB modules, respectively. From the images in columns three and four, one can observe that (1) when replacing the IB module by controlnet-pose the interactions between human and object will be unreasonable (indicated by blue circle); (2) when replacing CB module by controlnet-inpaint the identity of product object will change (indicated by yellow circle). These phenomenons demonstrate that IB and CB modules focus mostly on the interactions between human and object and the ID-consistency of product, respectively. Finally, our \emph{VirtualModel} can generate good images by using both IB and CB modules. 

Furthermore, we show more comparison examples in Figure \ref{fig_morecomp}. In summary, the OHG task cannot be easily achieved by merely merging multiple controlnets. It is essential to explicitly consider the interaction between humans and objects, as well as maintaining the ID-consistency of the products, as successfully accomplished by our \emph{VirtualModel}.

\begin{figure*}[!htb]
  \centering
  \includegraphics[width=1.0\linewidth]{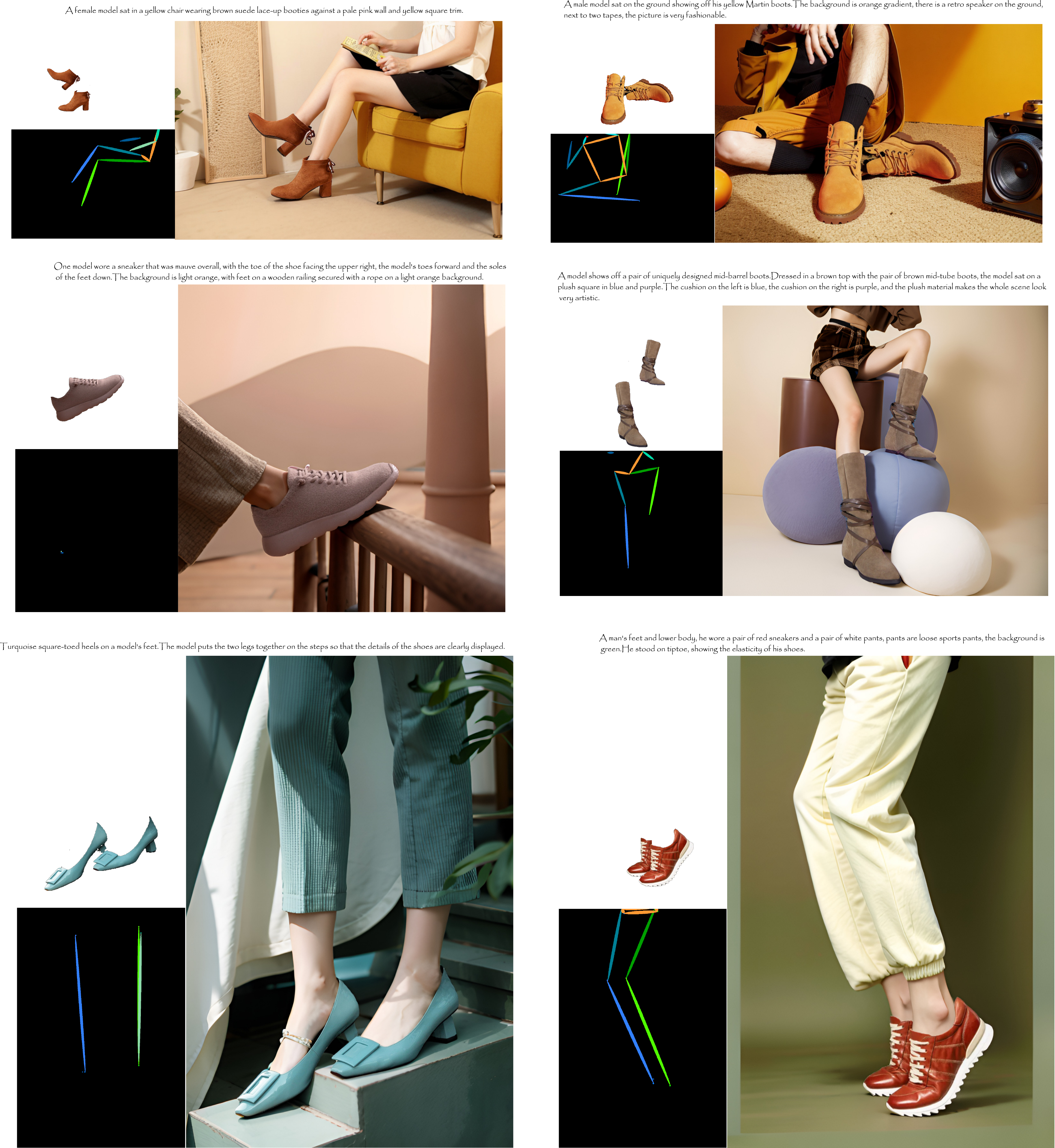}
  \caption{Additional qualitative results of \emph{VirtualModel} on \textbf{shoes}.}\label{fig_shoe}
\end{figure*}
\begin{figure*}[!htp]
  \centering
  \includegraphics[width=1.0\linewidth]{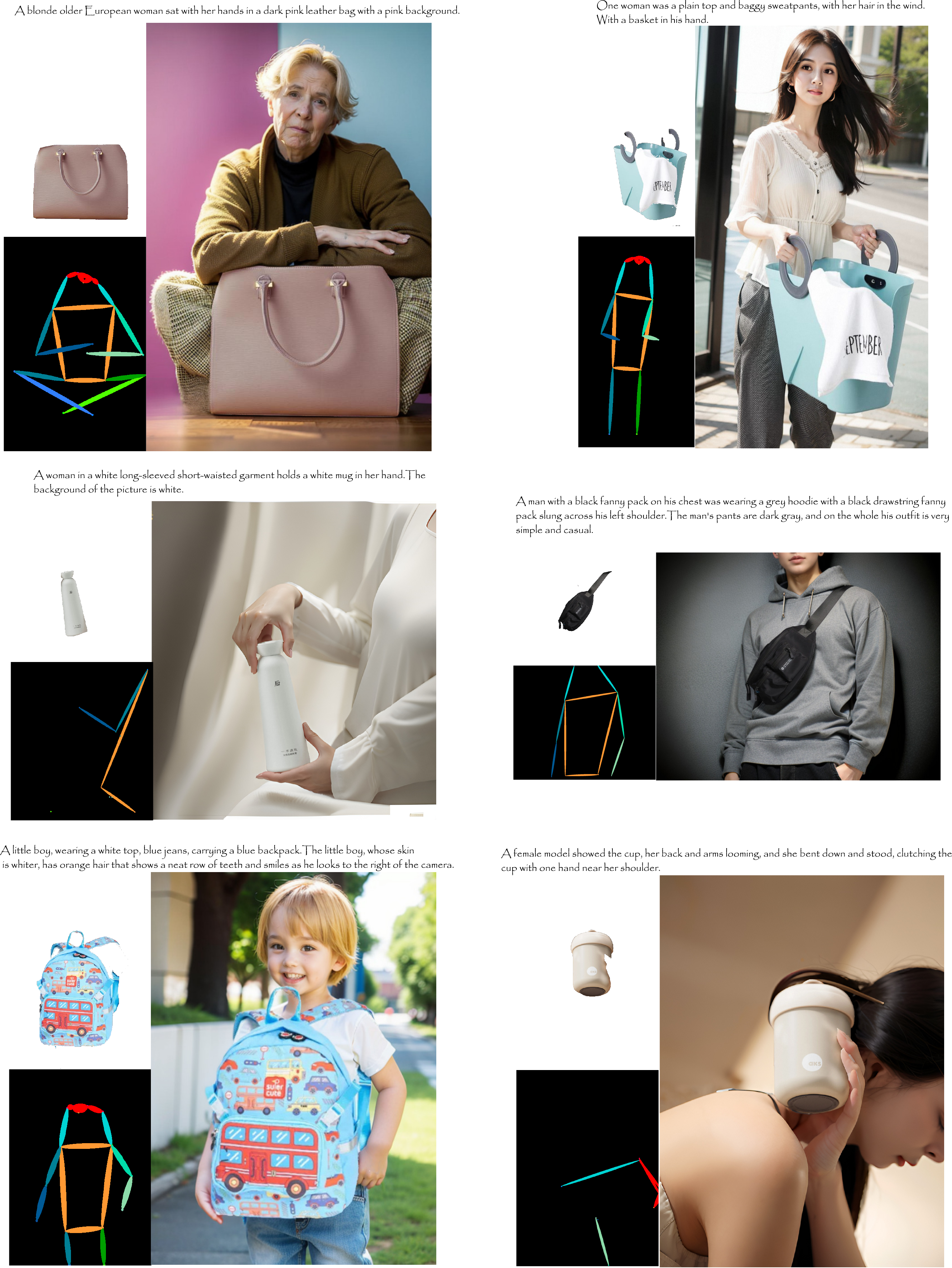}
  \caption{Additional qualitative results of \emph{VirtualModel} on \textbf{bags} and \textbf{bottles}.}\label{fig_bag}
\end{figure*}
\begin{figure*}[!htp]
  \centering
  \vspace{-1em}
  \includegraphics[width=0.95\linewidth]{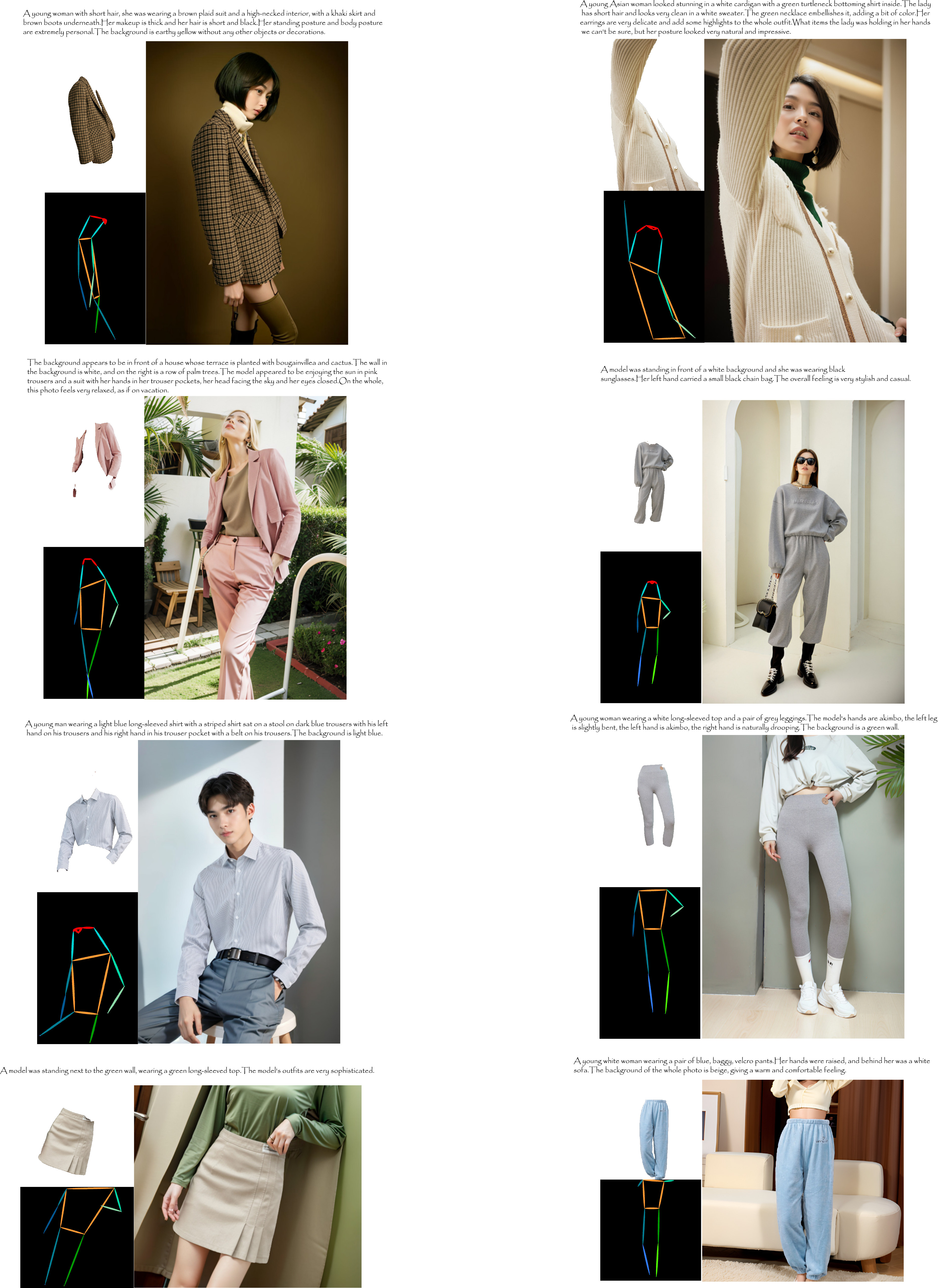}
  \caption{Additional qualitative results of \emph{VirtualModel} on \textbf{clothes}.}\label{fig_cloth}
  \vspace{-1em}
\end{figure*}
\begin{figure*}[!htp]
  \centering
  \includegraphics[width=0.9\linewidth]{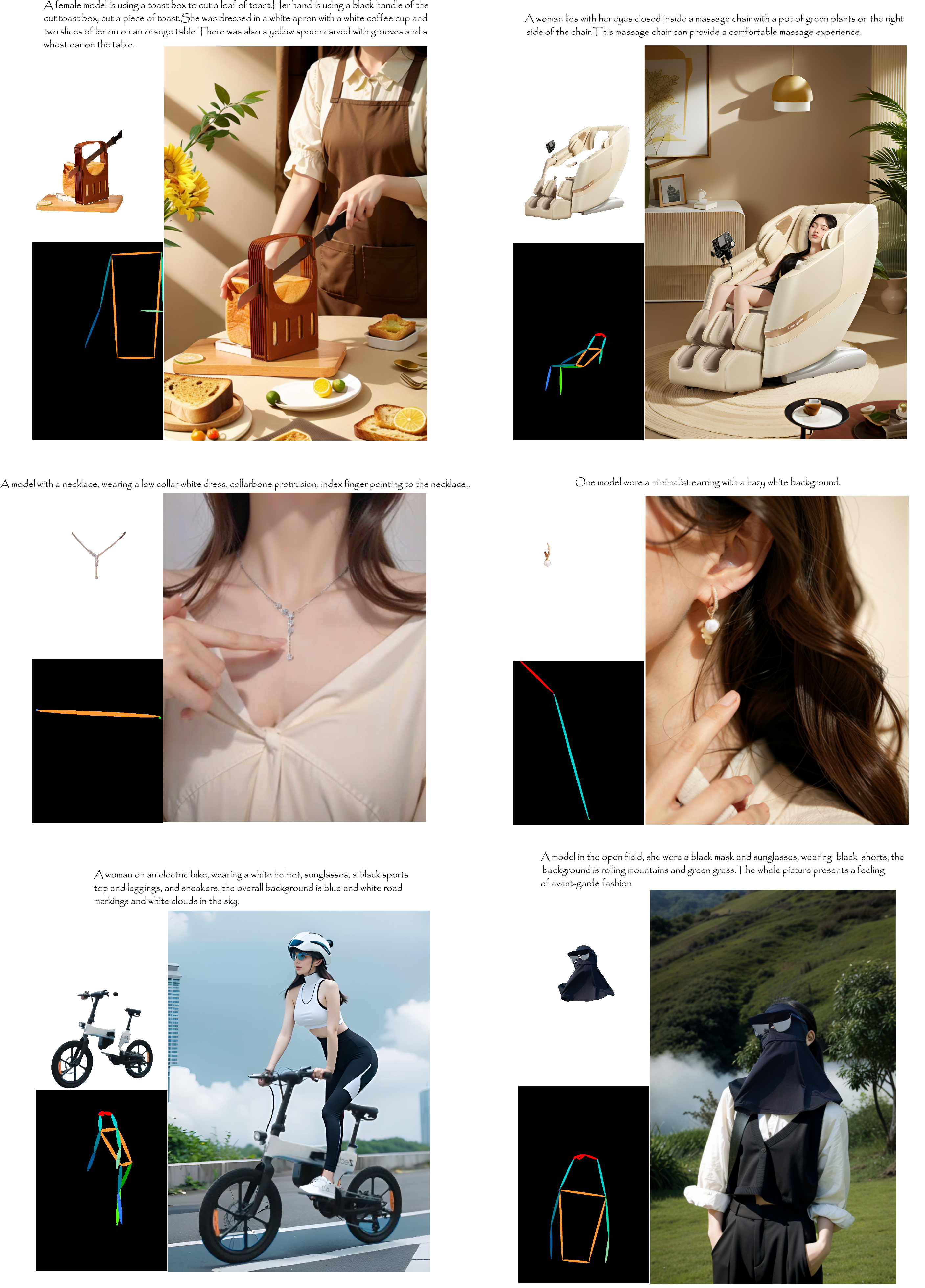}
  \caption{Additional qualitative results of \emph{VirtualModel} on other products.}\label{fig_other_products}
\end{figure*}

\textbf{3. Additional qualitative results of \emph{VirtualModel}}
In addition, we provide more example results of our \emph{VirtualModel}, including shoes, bags, bottles, clothes and other products, as in Figure \ref{fig_shoe},\ref{fig_bag},\ref{fig_cloth},\ref{fig_other_products}. From these figures, one can observe that our \emph{VirtualModel} supports for diverse products and interactions, and indeed is good for e-commerce marketing.

\begin{figure}[!hbp]
  \centering
  \includegraphics[width=0.9\linewidth]{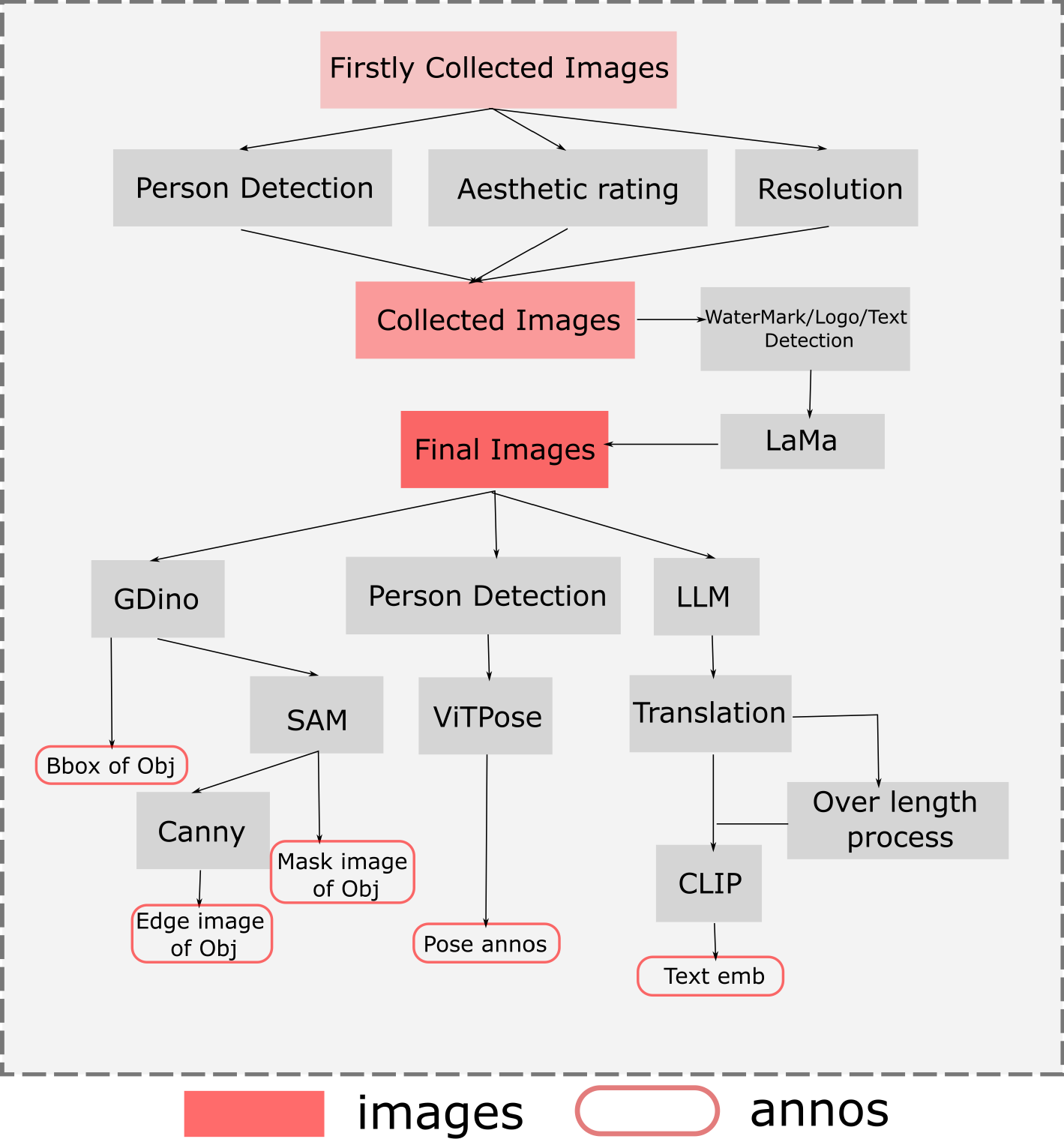}
  \caption{Pipeline of the construction of our HoIHuman Dataset.}\label{fig_hoi_pipeline}
\end{figure}
\section{More Details about HoIHuman Dataset}
The construction pipeline of our HoIHuman dataset is shown in Figure \ref{fig_ab_vm} and some details are listed as follows:

\textbf{Yolox person detection threshold }: 0.5

\textbf{Aesthetic rating }: \href{https://github.com/christophschuhmann/improved-aesthetic-predictor}{method github}, with threshold of 6

\textbf{Resolution filter}: images with the shortest side smaller than 256 are dropped.

\textbf{Person number per image}: only images featuring individuals numbered from 1 to 5 are retained.

\textbf{ViTPose model}: ViTPose-Huge wholebody checkpoint is used.

\textbf{Object edge detector}: canny with threshold 75 and 100

\textbf{GroundingDino text input}: "cosmetics, clothes, shoes, accessories, electronic products, bottle, cup, furniture, jacket, pants, dress, hat, glasses, coat, sneaker, phone, book"

\textbf{SAM}: input is the box information from groundingDino.

\textbf{OCR detection for watermark/logo/text:}  we use \href{https://www.modelscope.cn/models/damo/cv_convnextTiny_ocr-recognition-general_damo/summary}{modelscope api} for this detection.

\textbf{LaMa for image cleanup:} we use \href{https://www.modelscope.cn/models/damo/cv_fft_inpainting_lama/summary}{modelscope api}. And use its refinement strategy for our high-resolution images.

\textbf{Translation:} Since QWenVL focus on chinese, we use another chinese2english translation model (\href{https://www.modelscope.cn/models/damo/nlp\_csanmt\_translation\_en2zh/summary}{CSANMT}) for obtaining the corresponding English prompt.

\textbf{Prompt Length}: Since the length of prompt might be longer than 77, we split it to many sub-77-prompts and then encode them by CLIP model, and concat the output multiple 77-tokens together again.

\begin{figure}[!htp]
  \centering
  \includegraphics[width=1.0\linewidth]{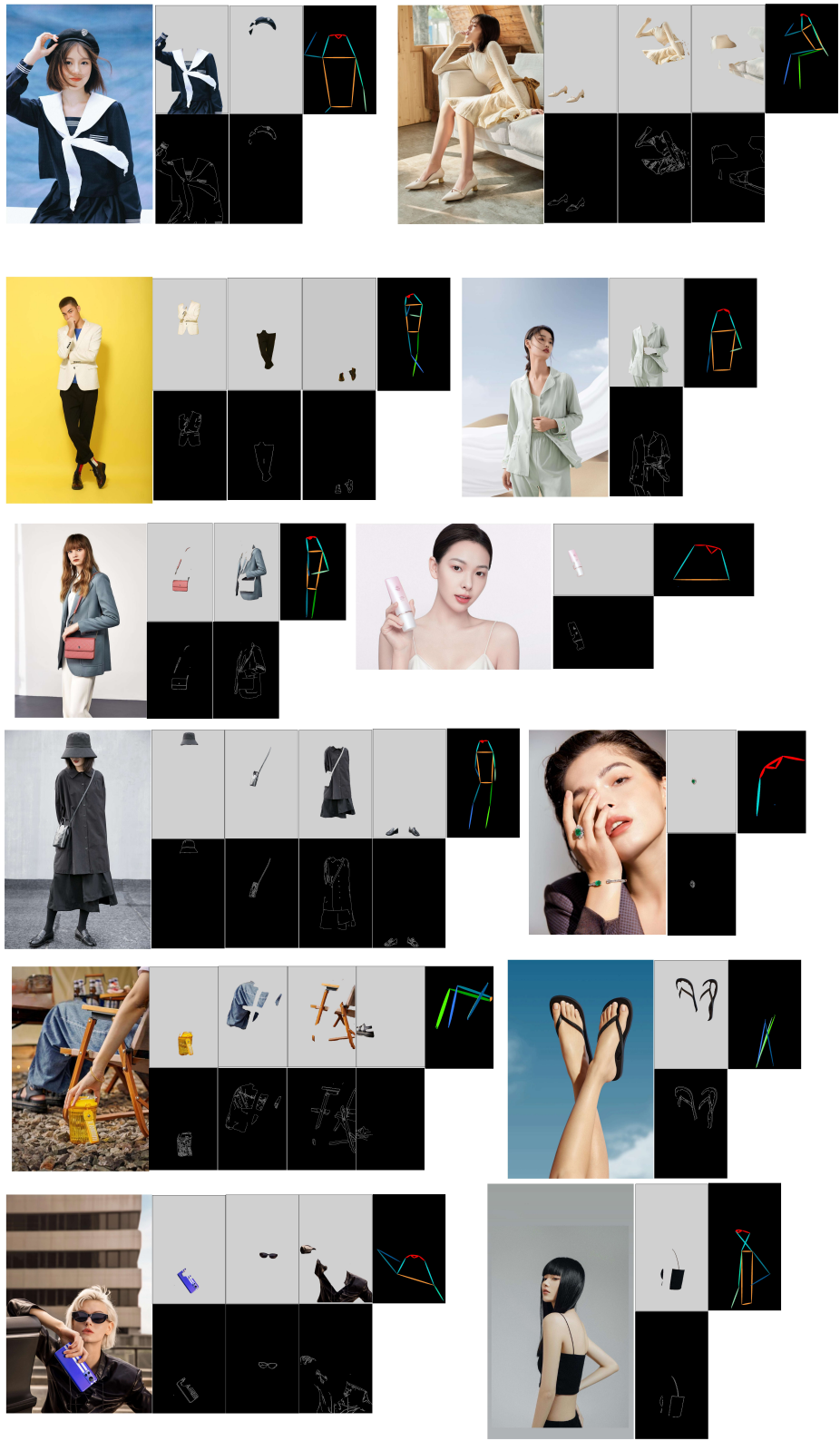}
  \caption{More example images from our HoIHuman dataset. We show the pose-skeleton images, object masks and the corresponding object edge images. \textbf{Best viewed with zoom-in}}\label{fig_hoi}
\end{figure}

Furthermore, to enhance the comprehension of our HoIHuman dataset, we have randomly chosen several examples and presented them in Figure \ref{fig_hoi}. It is evident that our HoIHuman dataset comprises high-quality images and annotations, all of which are tailored specifically for e-commerce marketing purposes.
\section{Details about IB and CB}
\textbf{Architecture details:} As described in the main paper, we employ a UNet architecture for our model, but with fewer channels in each layer, specifically 40$\%$ of the channels compared to the base UNet. The features $F_{IB},F_{CB},F_{H}$ that are passed between the IB (or CB) and the base UNet are encoded using 3x3 convolution layers initialized with zeros, ensuring the appropriate input and output channels.

\textbf{Training details:} Since each image might have multiple product objects, during training, we randomly choose one product object (and its corresponding edge image and visual embeddings) at each iteration.

%
%
\bibliographystyle{splncs04}
\bibliography{main}
\end{document}